\documentclass[final]{cvpr}

\usepackage{times}
\usepackage{epsfig}
\usepackage{graphicx}
\usepackage{amsmath}
\usepackage{amssymb}
\usepackage{float}
\usepackage{xcolor}
\usepackage{stfloats}
\usepackage{graphicx,float}
\usepackage{pifont}
\usepackage{enumitem}
\usepackage{booktabs}

\usepackage[pagebackref=true,breaklinks=true,colorlinks,bookmarks=false]{hyperref}

\def\cvprPaperID{20} 
\def\confYear{CVPR 2021}

\begin{document}

\title{Exploring Vision Transformers for Fine-grained Classification}

\author{Marcos V. Conde\\
Universidad de Valladolid\\
{\tt\small drmarcosv@protonmail.com}

\and

Kerem Turgutlu\\
University of San Francisco\\
{\tt\small kcturgutlu@dons.usfca.com}
}

\maketitle



\begin{abstract}

Existing computer vision research in categorization struggles with fine-grained attributes recognition
due to the inherently high intra-class variances and low inter-class variances.
SOTA methods tackle this challenge by locating the most informative image regions and rely
on them to classify the complete image. The most recent work, Vision Transformer (ViT),
shows its strong performance in both traditional and fine-grained classification tasks.
In this work, we propose a multi-stage ViT framework for fine-grained image classification tasks, which localizes the informative image regions without requiring architectural changes using the inherent multi-head self-attention mechanism. We also introduce attention-guided augmentations for improving the model's capabilities.
We demonstrate the value of our approach by experimenting with four popular fine-grained benchmarks: CUB-200-2011, Stanford Cars, Stanford Dogs, and FGVC7 Plant Pathology. We also prove our model's interpretability via qualitative results. 
See {\small{\url{https://github.com/mv-lab/ViT-FGVC8}}}.

%

\end{abstract}

\section{Introduction}
\label{introduction}

How to tell a dog’s breed, a car's brand, or a bird's species?. These are all challenging tasks even to the average human and usually require expertise. 
Fine-Grained Visual Classification (FGVC) aims to classify the sub-categories under coarse-grained large categories. Not only does it require to recognize a dog in the image but also correctly tell whether it is a Siberian Husky or an Alaskan Malamute.
FGVC is challenging because objects that belong to different categories might have similar characteristics, but differences between sub-categories might be remarkable (small inter-class variations and large intra-class variations). Because of these reasons, it is hard to obtain accurate classification results using classical Convolutional Neural Networks \cite{imagenet, resnet, inception, vgg, effnet}. 

\begin{figure}[htp!]
    \begin{center}
    \includegraphics[width=0.9\linewidth]{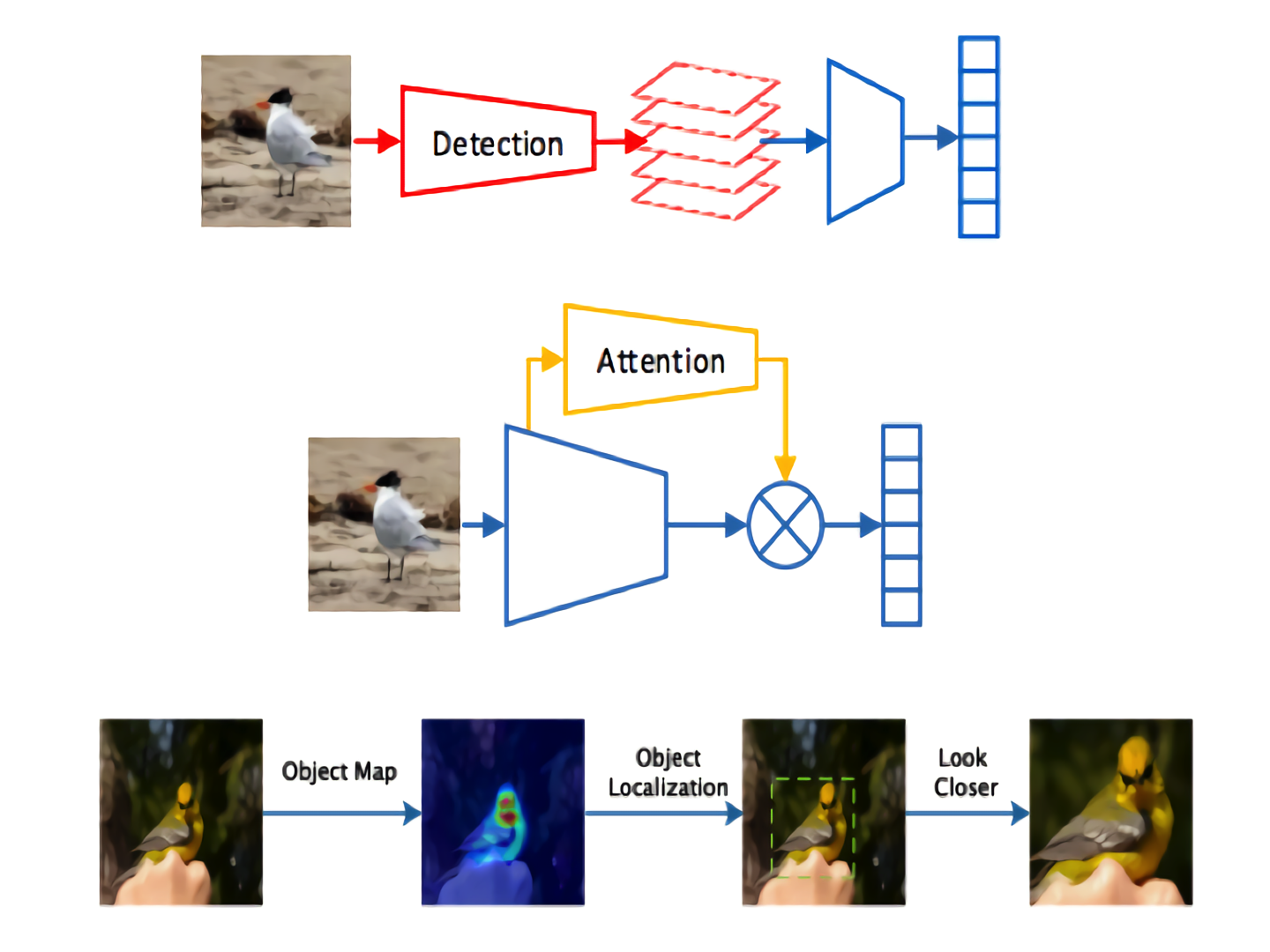}
    \end{center}
    \caption{SOTA methods based on localization of discriminative regions. Diagrams from Chen et.al. \cite{Chen_2019_CVPR_destruction} and Hu et.al. \cite{hu2019better_fgvc}.}
    \label{fig:teaser}
\end{figure}

Recent work shows the key step of FGVC is identifying and extracting more informative regions and features in an image \cite{ Lam_2017_CVPR_fgvc, Chen_2019_CVPR_fgvc, hu2019better_fgvc, zhang2020multibranch, fgvc_cvpr2020_details}. This can be done using a backbone network (i.e. ResNet \cite{resnet}) to extract features of the image and selected regions. However, this strategy inevitably leads to complex pipelines, sometimes not fully differentiable, and pushes the proposed regions to contain most of the parts of the objects.
Moreover, labeling fine-grained categories is an expensive and time-consuming process that requires expertise in a specialized domain. Therefore, FGVC datasets \cite{data_KhoslaYaoJayadevaprakashFeiFei_FGVC2011, data_WahCUB_200_2011, data_maji2013finegrained} often have limited training data. \vspace{\baselineskip}

\textbf{Our main contributions are:} (i) An interpretable multi-stage model based on Vision Transformers \cite{vit} and detection-based FGVC methods, that allows to localize and recognize informative regions in the image using the inherent multi-head attention mechanism.
(ii) Augmentations based on attention flow with respect to the input tokens that lead to forcing the image to learn and extract a broader set of fine-grained features. (iii) Explore potential of Visual Transformers for fine-grained classification and achieve state-of-the-art performance.

\section{Approach}
\label{approach}

\begin{figure*}[ht!]
    \begin{center}
    \includegraphics[width=0.8\linewidth]{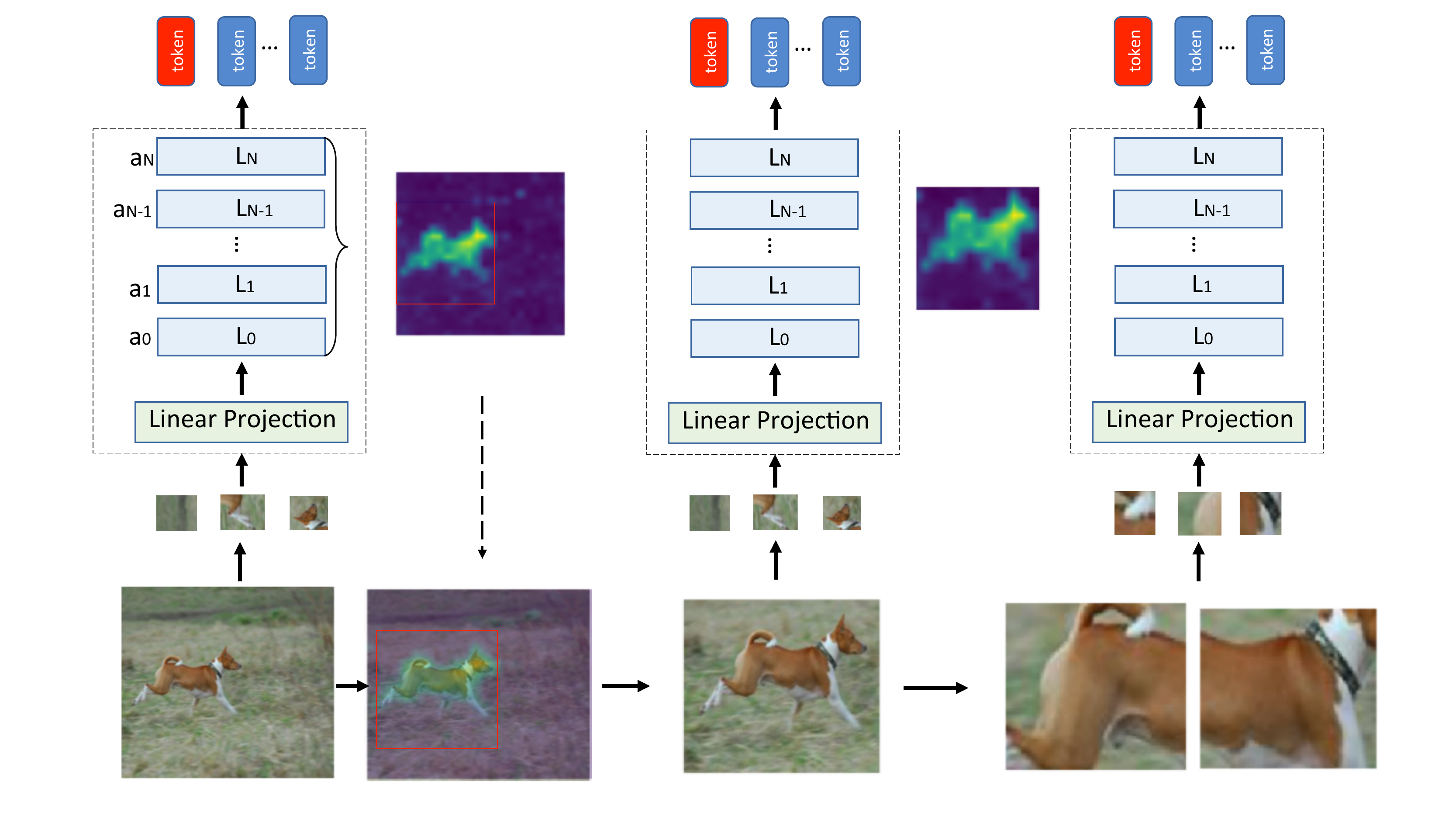}
    \put(-342,0){ (a) }
    \put(-195,0){ (b) }
    \put(-80,0){ (c) }
    \put(-287,180){ \scriptsize Attention Map }
    \put(-285,120){ \scriptsize Localization }
    \put(-305,210){ Tokens }
    \put(-385,100){ \textbf {\textit{E} }}
    \put(-235,100){ \textbf {\textit{E} }}
    \put(-120,100){ \textbf {\textit{E} }}
    \end{center}
    \caption{Summary of our approach as explained in Section \ref{approach}, there are three stages a,b,c, and the ViT body is shared across them. We also show the corresponding self-attention map at stage (a) with its detected bounding box in red color.}
    \label{fig:main}
\end{figure*}
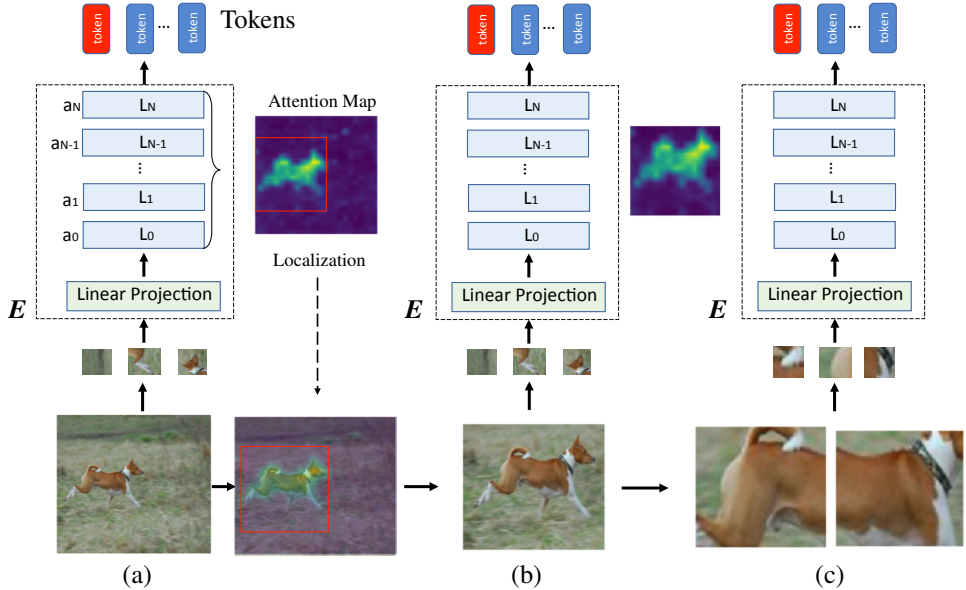

Our Multi-stage Multi-scale model, inspired by MMAL-Net \cite{zhang2020multibranch} can be seen in Figure \ref{fig:main}. 

We use a Vision Transformer (ViT) \cite{vit} for encoding images into feature representation for further downstream fine-grained classification task and use the inherent multi-head self-attention mechanism to extract attention maps with respect to the original input image using the attention flow method from Abnar et.al.  \cite{abnar2020quantifying}. Inherent multi-head self-attention allows us to simultaneously extract features and localize important regions in a single forward pass without additional memory requirements or architectural changes. In our experiments we used ViT-B/16 as our encoder \textit{\textbf{E}} and a 2 layer \textit{\textbf{MLP}} with \textit{ReLU} activation in between for the classification task. For classification, \textit{CLS} token, colored in red in Figure \ref{fig:main}, is fed to \textit{\textbf{MLP}}.
Our method composes of 3 stages in sequence:
\begin{enumerate}[label=\alph*]
    \item \textbf{Main stage}: The main stage uses encoder \textbf{\textit{E}} to extract attention maps. A bounding box is generated using the object localization method. In this stage, the full image feature is used for the downstream classification task. Furthermore, the object and its corresponding attention map are cropped from a higher resolution version of the full image and passed down to the object stage.
    \item \textbf{Object stage}: In the object stage, cropped image feature is used for the downstream classification task. 
    \item \textbf{Parts stage}: Cropped attention map from previous stages is used to extract fine-grained crops from the localized object image using the discriminative regions localization method. In this stage, crops are used for the downstream classification task as independent inputs.
\end{enumerate}
Note that all parameters are shared across stages as in \cite{zhang2020multibranch}, and they are optimized simultaneously using the loss function $\mathcal{L}_{joint}$ defined in Equation \ref{loss}, based on the Cross-Entropy loss denoted as $\mathcal{L}$, where $\mathcal{L}_a$, $\mathcal{L}_b$, $\mathcal{L}_c$ correspond to the three stages explained in this Section and shown in Figure \ref{fig:main}. Note that for stage c, $\mathcal{L}_c$ is the aggregation of losses from the respective $N$ ${c}_i$ crops.
\begin{equation}
\begin{aligned}[b]
    \mathcal{L}_{joint} = \mathcal{L}_a + \mathcal{L}_b + \sum_{i=1}^N \mathcal{L}_{c_i}
\end{aligned}
\label{loss}
\end{equation}
%
The Attention Map \textit{AM} is obtained by recursively multiplying the self-attention weights $a_{0} \dots a_{N}$ from each transformer layer $L_{0} \dots L_{N}$, this recovers more information than using only the last layer.

\subsection{Object Localization}
\label{localization}

The attention map of the full image is binarized using a histogram-based thresholding algorithm. Then, a grayscale morphological closing is applied on this binary map. Finally, the bounding box that covers the binary foreground with the maximum attention value is used for object cropping. During our experiments, we found that mean thresholding \cite{GLASBEY1993532} to be working well.

\subsection{Discriminative Regions Localization}
\label{parts-localization}

Object part regions (c) are extracted using the attention map of the object after cropping it from the full attention map as seen in (b) from Figure \ref{fig:main}. The strength of attention in local areas can be an indicator of the importance of these regions captured by the model. Therefore, we search for such regions which cover the most attention and at the same time are not very overlapped. This allows us to find non-overlapping object part crops.
We calculate scores for each object part region candidate in the localized object's attention map. These scores are calculated using Average Pooling (AP) layer with a kernel size of $H \times W$ and stride $S = 1$. Stride $S = 1$ is used to search all possible regions but its value can be increased to reduce the computational overhead. We empirically found that Max Pooling layer leads to worse performance as it is too localized. However, generalized-mean (GeM) pooling can also be used \cite{radenovic2018finetuning}. Kernel size of $H \times W$ is a hyper-parameter that can be tuned based on the dataset. In our experiments, we used 0.3 of the cropped object area, which roughly accounts for a kernel size of 112x112 when object attention maps are resized to 384 px.
After the scoring, we apply Non-Maximum Suppression (NMS) similar to other well know detectors \cite{girshick2015fast} to get the top non-overlapping regions. We used the top 2 object regions after NMS.

\subsection{Attention-based Augmentations}
\label{augs}
\begin{figure}[H]
    \begin{center}
    \includegraphics[width=\linewidth]{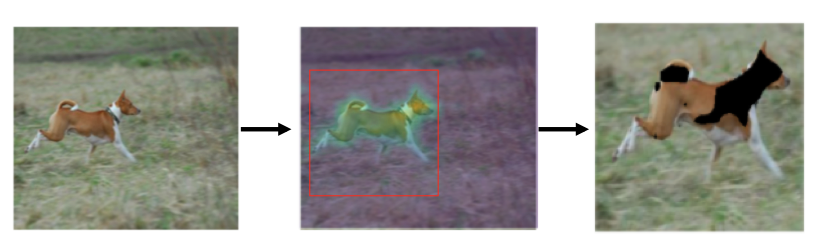}
    \end{center}
    \caption{Attention-based cropping and erasing (see Section \ref{augs}).
    }
    \label{fig:augs}
\end{figure}

Data augmentation allows to increase the variety of training data, prevent overfitting and improve the performance of deep models. However, in FGVC, random data augmentation, such as random image cropping, is low-efficiency and might remove informative parts.
Inspired by WS-DAN \cite{hu2019better_fgvc}, we augment the image using the learned self-attention map, which represents the importance of image parts (tokens) based on the model. As shown in Figure \ref{fig:augs}, we perform cropping and erasing. Removing currently attended pixels enforces the model to consider other discriminative regions and allows feature diversity during training. In a given batch, we randomly apply random erasing on an image with a probability \textit{P} by erasing all the pixels that have an attention value higher than the threshold \textit{T}.
Particularly \textit{RandomErasing} \cite{zhong2017random}, as a destructive augmentation, has proven to improve accuracy performance on fine-grained classification \cite{Chen_2019_CVPR_fgvc, hu2019better_fgvc}.

\section{Experiments}
In this section we report the results for each benchmark dataset. In each experiment we used the same setup for training. 
We use augmentations proposed in Section \ref{augs} together with standard augmentations (\textit{e.g.} horizontal flip, additive noise, etc.)
Full images (stage a) are resized to 384 px resolution in main stage, objects (stage b) and crops (stage c) are cropped from a higher resolution and then resized to 384 px before the forward pass to the next stage.

\begin{table}[H]
    \begin{center}
    \begin{tabular}{c c c c}
    \toprule
    Datasets & Category & Training & Testing \\
    \midrule
    CUB-Birds \cite{data_WahCUB_200_2011} & 200 & 5994 & 5794 \\
    Stanford Dogs \cite{data_KhoslaYaoJayadevaprakashFeiFei_FGVC2011} & 120 & 12000 & 8580 \\
    Stanford Cars \cite{KrauseStarkDengFei-Fei_3DRR2013} & 196 & 8144 & 8041 \\
    Plant Pathology \cite{mwebaze2019icassava} & 5 & 9436 & 12595 \\ 
    \bottomrule
    \end{tabular}
    \end{center}
    \caption{Summary of Datasets.}
    \label{tab:datasets}
\end{table}


\begin{table}[H]
    \begin{center}
    \begin{tabular}{c c c}
    \toprule
    Method & Backbone & Accuracy(\%) \\
    \midrule
    MaxEnt \cite{maxent} & DenseNet-161 & 83.6 \\
    FDL [REF] & DenseNet-161 & 84.9 \\
    RA-CNN \cite{Fu_2017_CVPR_racnn} & VGG-19 & 87.3 \\
    Cross-X \cite{luo2019crossx} & ResNet-50 & 88.9 \\
    API-Net \cite{zhuang2020learning_apinet} & ResNet-101 & 90.3 \\
    ViT \cite{vit} & \text{ViT-B/16} & 91.7 \\
    \midrule
    Ours & \text{ViT-B/16}  & \textbf{93.2} \\
    \bottomrule
    \end{tabular}
    \end{center}
    \caption{Comparison of different methods on Stanford Dogs and ablation of our method.}
    \label{table:dogs}
\end{table}


\begin{table}[H]
    \begin{center}
    \begin{tabular}{c c c c}
    \toprule
    Method & Backbone & CUB & Cars \\
    \midrule

    VGG-16 \cite{vgg} &  VGG-16& 77.8 & 85.7 \\
    ResNet-101 \cite{resnet} &  ResNet-101 & 83.5 &  91.2 \\
    Inception-V3 \cite{inception} & Inception-V3 & 83.7 &  90.8 \\
    \midrule
    RA-CNN \cite{Fu_2017_CVPR_racnn}& VGG-19 & 85.3 & 92.5 \\
    MaxEnt \cite{maxent}& DenseNet-161 & 86.6 & 93.0 \\
    Cross-X \cite{luo2019crossx} & ResNet-50 & 87.7 & 94.6 \\
    DCL \cite{Chen_2019_CVPR_destruction} & ResNet-50 & 87.8 & 94.5 \\
    API-Net \cite{zhuang2020learning_apinet}& DenseNet-161 & 90.0 & 95.3 \\
    WS-DAN \cite{hu2019better_fgvc} & Inception v3 & 89.4 & 94.5 \\
    MMAL-Net \cite{zhang2020multibranch} & ResNet-50 & 89.6 & \textbf{95.0} \\
    ViT \cite{vit} & \text{ViT-B/16} & 89.4 & 92.8 \\
    \midrule
    Ours & \text{ViT-B/16} & \textbf{91.0} & 94.3 \\
    \bottomrule
    \end{tabular}
    \end{center}
    \caption{Comparison with state-of-the-art methods on CUB-200-2011 and Stanford Cars.}
    \label{table:cars-cub}
\end{table}


\begin{table}[H]
    \begin{center}
    \begin{tabular}{c c c}
    \toprule
    Method & Backbone & Accuracy(\%) \\
    \midrule
    ResNet \cite{resnet} & 50 & 89.2 \\
    EfficienNet\cite{effnet} & B0 & 90.1 \\
    EfficienNet\cite{effnet} & B2 & 90.4 \\
    ViT \cite{vit} & \text{ViT-B/16} & 91.7 \\
    \midrule
    Ours & \text{ViT-B/16} & \textbf{92.4} \\
    \bottomrule
    \end{tabular}
    \end{center}
    \caption{Method comparison on FGVC7 Plant Pathology.}
    \label{table:plants}
\end{table}

\subsection{Results}
The reported results in Tables \ref{table:dogs},\ref{table:cars-cub},\ref{table:plants} show that Vision Transformers have great potential on FGVC. Moreover, our experiments show that attention-driven augmentations and important regions detection help to improve their performance on fine-grained classification, achieving state-of-the-art performance in Stanford Dogs \cite{data_KhoslaYaoJayadevaprakashFeiFei_FGVC2011} and CUB-200-2011 \cite{data_WahCUB_200_2011}.
However, we must expose the \textbf{limitations} of the proposed multi-branch multi-scale architecture: (i) the selection of the region of interest (ROI) based on the attention map is not fully differentiable, and thus, the model is not complete end-to-end trainable, it requires to train it in a sequential (multi-stage) way. (ii) ViT-based models require important computational power.

\begin{figure*}[ht!]
    \begin{center}
    \includegraphics[width=\linewidth, scale=1.5]{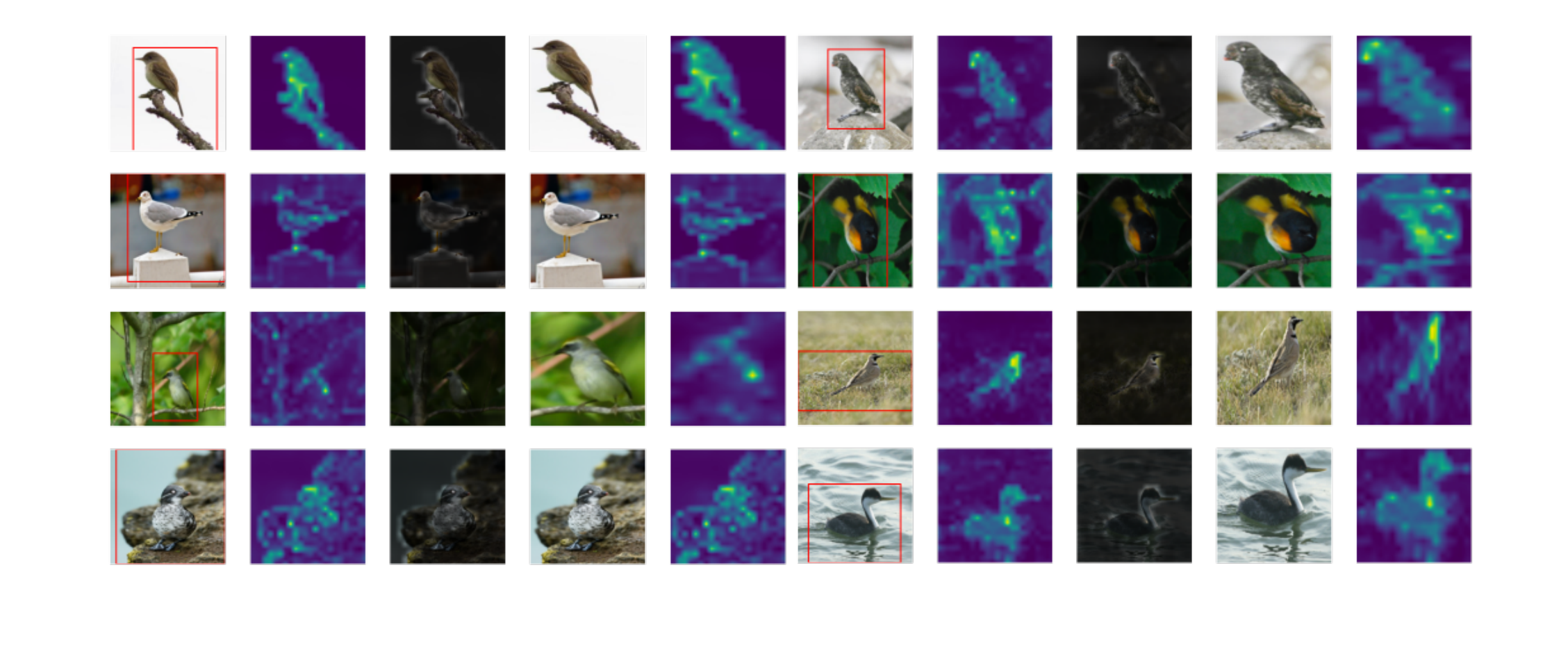}
    \put(-450,10){(a)}
    \put(-405,10){(b)}
    \put(-360,10){(c)}
    \put(-318,10){(d)}
    \put(-272,10){(e)}
    \put(-230,10){(a)}
    \put(-185,10){(b)}
    \put(-145,10){(c)}
    \put(-100,10){(d)}
    \put(-55,10){(e)}
    \end{center}
    \caption{Results for CUB-200-2011 Dataset \cite{data_WahCUB_200_2011}. We show (a) input image, (b) attention map, (c) image after applying the global attention map, highlighting informative regions, (d) attention crop, from the predicted red bounding box and (e) crop's attention map. These qualitative visualizations prove the effectiveness and the interpretability of our method.
    8 complete results. Best viewed in color.}
    \label{fig:results_birds}
\end{figure*}

\begin{figure}[ht!]
    \begin{center}
    \includegraphics[trim={0 0.8cm 0 0},clip,width=\linewidth]{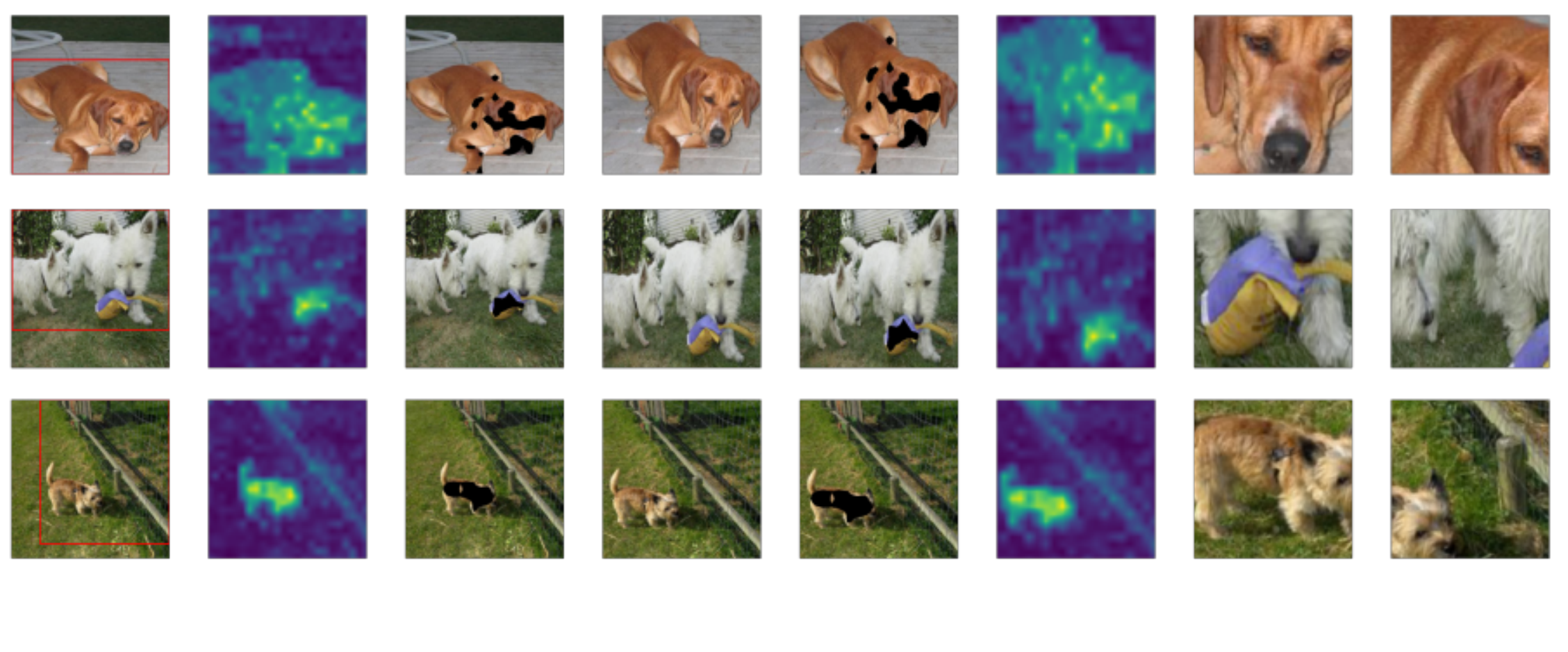}
    \end{center}
    \caption{Qualitative results for Stanford Dogs \cite{data_KhoslaYaoJayadevaprakashFeiFei_FGVC2011}. We show 3 samples, from left to right: the input image with the predicted bounding box (red color), the global attention map,
    attention-based erasing augmentation (\textit{RandomErasing}), selected crops and their attention map, the top-2 most informative regions in the image based on crop's attention (Section \ref{parts-localization}). Best viewed in color.}
    \label{fig:results_dogs}
\end{figure}


\section{Conclusion}

In this paper, we propose a multi-stage multi-scale fine-grained visual classification framework based on ViT. The multi-head self-attention mechanism can capture discriminative image features from multiple diverse local regions. We aim to exploit this property in our framework, and by doing it we achieve state-of-the-art results on popular fine-grained benchmarks. We also employ different attention-guided augmentations to improve our model's generalization capability by enforcing the model to learn more diverse discriminative features.
As future work, we aim towards making our framework end-to-end trainable and improve this approach by exploring detection transformers.


{\small
\bibliographystyle{ieee_fullname}
\bibliography{egbib}
}


\end{document}